\documentclass[journal,twoside,web]{IEEEtran}
\usepackage{rotating}
\usepackage{amsmath,amssymb,amsfonts}
\usepackage{algorithmic}
\usepackage{multirow}
\usepackage{graphicx}
\usepackage{textcomp}
\usepackage{orcidlink}
\usepackage[nolist]{acronym}
\usepackage{booktabs}
\usepackage{array,tabularx}
\newcolumntype{P}[1]{>{\raggedright\arraybackslash}p{#1}}
\usepackage[
    backend=biber,
    style=ieee,
    citestyle=numeric-comp,
    maxnames=4,
    minnames=1,
    url=false,
    doi=false,
    isbn=false,
    eprint=false
]{biblatex}

\AtEveryBibitem{%
  \clearfield{month}%
  \clearfield{day}%
  \clearfield{date}%
  \clearlist{editor}%
}
\AtEveryBibitem{%
  \clearfield{volume}%
  \clearfield{number}%
}

\def\BibTeX{{\rm B\kern-.05em{\sc i\kern-.025em b}\kern-.08em
    T\kern-.1667em\lower.7ex\hbox{E}\kern-.125emX}}
\begin{document}
\title{Bayesian Optimization for Design Parameters of 3D Image Data Analysis}
\author{David Exler\orcidlink{0009-0005-4116-5287}, 
Joaquin Eduardo Urrutia Gómez \orcidlink{0000-0001-5411-9689},
Martin Krüger \orcidlink{0009-0006-3545-4477},
Maike Schliephake, \\
John Jbeily,
Mario Vitacolonna\orcidlink{0000-0002-8599-1696}, 
Rüdiger Rudolf\orcidlink{0000-0002-0833-1053}, and 
Markus Reischl \orcidlink{0000-0002-7780-6374}
\thanks{This work has been submitted to the IEEE for possible publication.
Copyright may be transferred without notice, after which this version may no longer be accessible.}
\thanks{R. Rudolf was supported by DFG grant INST874/9-1.}
\thanks{D. Exler (e-mail: david.exler@kit.edu), J. E. Urrutia Gómez (e-mail: joaquin.urrutia-gomez@kit.edu), M. Krüger (e-mail: martin.krueger2@kit.edu), and M. Reischl (e-mail: markus.reischl@kit.edu) are with the Institute for Automation and Applied Informatics, Karlsruhe Institute of Technology, Hermann-von-Helmholtz-Platz 1, 76344 Eggenstein-Leopoldshafen, Germany.}
\thanks{J. Jbeily (e-mail: j.jbeily@doktoranden.th-mannheim.), M. Vitacolonna (e-mail: m.vitacolonna@hs-mannheim.de), and R. Rudolf (e-mail: r.rudolf@hs-mannheim.de) are with the CeMOS Research and Transfer Center, Technische Hochschule Mannheim, Paul-Wittsack-Straße 10, 68163 Mannheim, Germany.}
\thanks{M. Schliephake (e-mail: maike.schliephake@student.kit.edu) is with the Institute of Biological and Chemical Systems, Karlsruhe Institute of Technology, Hermann-von-Helmholtz-Platz 1, 76344 Eggenstein-Leopoldshafen, Germany.}}

\maketitle

\begin{abstract}
Deep learning-based segmentation and classification are crucial to large-scale biomedical imaging, particularly for 3D data, where manual analysis is impractical.
Although many methods exist, selecting suitable models and tuning parameters remains a major bottleneck in practice.
Hence, we introduce the 3D data Analysis Optimization Pipeline, a method designed to facilitate the design and parameterization of segmentation and classification using two Bayesian Optimization stages.
First, the pipeline selects a segmentation model and optimizes postprocessing parameters using a domain-adapted syntactic benchmark dataset.
To ensure a concise evaluation of segmentation performance, we introduce a segmentation quality metric that serves as the objective function.
Second, the pipeline optimizes design choices of a classifier, such as encoder and classifier head architectures, incorporation of prior knowledge, and pretraining strategies.
To reduce manual annotation effort, this stage includes an assisted class-annotation workflow that extracts predicted instances from the segmentation results and sequentially presents them to the operator, eliminating the need for manual tracking.
In four case studies, the 3D data Analysis Optimization Pipeline efficiently identifies effective model and parameter configurations for individual datasets.
\end{abstract}

\section{Introduction}
\label{sec:introduction}
\IEEEPARstart{M}{icroscopy} is a fundamental tool to observe, quantify, and interpret cell and tissue structures.
Recently, 3D data analysis has gained increasing relevance in cellular research because it captures spatial context lost in 2D projections \cite{ghose_3d_2023, du_spatial_2025}.
Modern devices can acquire hundreds to thousands of 3D images per sample, generating datasets that are nearly impossible to analyze manually.
Segmentation and classification are, therefore, key steps in automatically extracting interpretable information \cite{minaee_image_2022}.\\
Numerous methods have been proposed for segmentation and classification of 3D biomedical data \cite{gertych_rapid_2016, ling_mtanet_2024, shaker_unetr_2024}.
Most segmentation approaches are based on the U-Net architecture \cite{ronneberger_u-net_2015}, while classification models typically combine a \ac{cnn} or Vision Transformer encoder with a classifier head \cite{boser_training_1992, yarowsky_unsupervised_1995}.
Several pretrained segmentation models for 3D biomedical images are available and can be adapted to new datasets \cite{archit_segment_2025, israel_cellsam_2025, wolny_accurate_2020, schwartz_caliban_2024, pachitariu_cellpose-sam_2025}.
Postprocessing techniques such as morphological operations, watershed-based separation, or graph-based merging can further improve segmentation quality \cite{soille_automated_1990, meyer_color_1992, chang_region-based_2004, belaid_image_2009, neubert_compact_2014}.\\
For classification, encoder backbones pretrained on large-scale datasets like ImageNet are available \cite{russakovsky_imagenet_2015, he_deep_2016, liu_swin_2022, tan_efficientnetv2_2021, liu_convnet_2022}, but adapting them typically requires additional training.
Semi-supervised annotation methods, such as label spreading, can enhance datasets with limited annotations \cite{zhou_learning_2003}.
Classifier heads must be tailored to the downstream task, often incorporating explicit 3D operations such as 3D \acp{cnn} or self-attention mechanisms \cite{wang_non-local_2018, liu_group_2021}.
Overall, segmentation and classification methods require dataset-specific adaptation to achieve optimal performance, which is challenging due to the limited availability of annotated 3D data \cite{nikolenko_synthetic_2021}.\\
To mitigate this limitation, synthetic 3D data generation and domain adaptation methods have been proposed, reducing the need for labor-intensive manual annotation \cite{dunn_deepsynth_2019, fu_three_2018, eschweiler_3d_2021, bruch_improving_2025}.
Nevertheless, optimizing analysis pipelines for specific 3D tasks remains difficult.
Naive parameter tuning approaches such as grid or random search are computationally inefficient, whereas \ac{bo} enables efficient optimization of expensive black-box functions by modelling uncertainty with probabilistic surrogate models \cite{ohagan_curve_1978, jones_lipschitzian_1993}.\\
We therefore introduce the \acf{pipe}, a framework for efficient automated optimization of 3D image analysis pipelines.
The \ac{pipe} guides researchers from raw data to analysis by identifying dataset characteristics relevant to deep learning performance and optimizing segmentation and classification using two \ac{bo} processes.
The first process compares segmentation models and optimizes postprocessing parameters on a domain-adapted synthetic benchmark dataset, supported by a newly introduced quality criterion that distinguishes different segmentation error types.
Using the optimized segmentation to facilitate 3D instance annotation, the second \ac{bo} process optimizes classifier design parameters, including pretraining strategies, encoder and classifier head architectures, and the integration of prior knowledge.
We demonstrate the applicability of the \ac{pipe} on four cellular 3D microscopy datasets.
First, we analyze nuclei in \textit{in vitro} Myotube cultures \cite{couturier_aberrant_2024}.
Second, we apply the \ac{pipe} to an additional cellular core-shell dataset.
Finally, we perform segmentation optimization on two datasets from the \ac{ctc} \cite{maska_cell_2023}.
Across datasets with differing optical properties, the results show that optimal design choices and parameters are highly data-dependent.
The main contributions of this work are:
\begin{itemize}
    \item an experimentally validated workflow for automated optimization of 3D image analysis pipelines,
    \item a Bayesian Optimization process to adapt conceptual parameters of instance segmentation inference to a specific image data domain without computationally expensive model retraining, 
    \item a new metric to detect interpretable segmentation errors, and
    \item a Bayesian Optimization process to fine-tune design parameters of a classifier. 
\end{itemize}

\section{Methods}

\subsection{Overview}
\begin{figure*}
    \centering
    \includegraphics[width=0.99\linewidth]{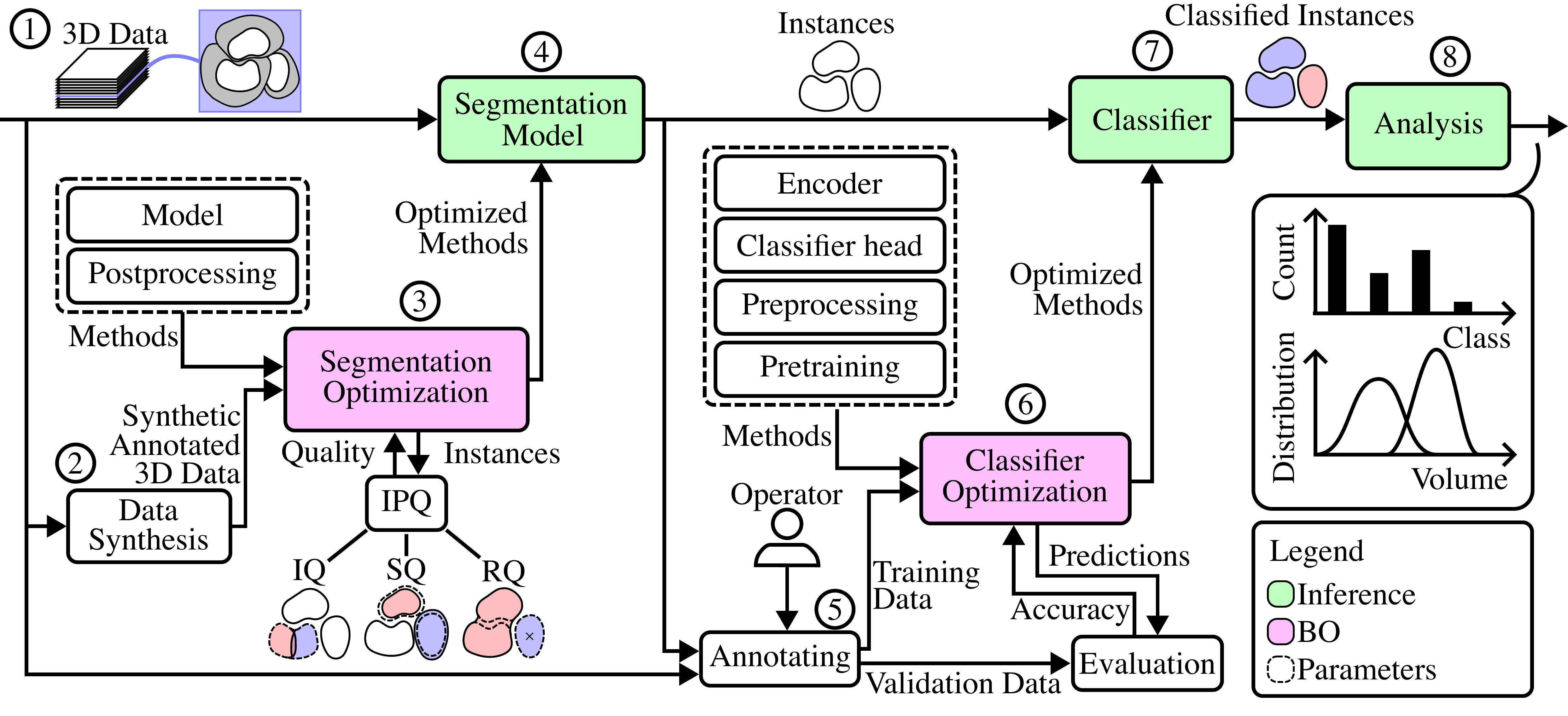}
    \caption{Progression of the introduced pipeline.
    3D data (1) is used for data synthesis (2), combining instance simulation and domain adaptation.
    The synthetic data serves as input for segmentation optimization (3), where a pretrained model and postprocessing parameters are optimized with respect to the \ac{ipq} metric.
    The optimized segmentation model (4) predicts instances, which are annotated by an operator (5).
    Classifier optimization (6) selects encoder and classifier head architectures, preprocessing, and pretraining strategies based on validation accuracy.
    The optimized classifier (7) predicts class labels, enabling the final analysis (8), which yields statistics such as class counts and instance volume distributions.}
    \label{fig:graph_abstract}
\end{figure*}

To mitigate the effort of manual 3D image data analysis pipeline design, we introduce the \ac{pipe}, a framework that optimizes the design of automated 3D data analysis pipelines using two \ac{bo} processes.\\
Fig.~\ref{fig:graph_abstract} illustrates the workflow of the \ac{pipe}.
Due to the limited availability of annotated 3D datasets, the pipeline begins with \textit{data synthesis} (2), generating a segmentation benchmark from the input \textit{3D data} (1).
A \textit{segmentation optimization} (3) then selects a pretrained segmentation model and optimizes postprocessing parameters using the newly introduced \ac{ipq} metric.
The resulting \textit{segmentation model} (4) predicts instances on the input data.
To reduce the effort of annotating 3D data for classification, the predicted instances are integrated into an assisted annotation workflow, enabling efficient \textit{annotation} (5).
Subsequently, a \textit{classifier optimization} (6) identifies optimal classifier design parameters, including encoder and classifier head architectures, preprocessing, and pretraining strategies, using validation accuracy as the objective.
The optimized \textit{classifier} (7) produces class predictions for each instance, enabling the final \textit{analysis} (8).
\noindent
In contrast to recent research that optimizes training hyperparameters, the \ac{pipe} focuses on optimizing conceptual design parameters \cite{ramalakshmi_u-net-based_2025, douglas_automatic_2025}.

\subsection{Segmentation Optimization}

\subsubsection{Objective: Injective Panoptic Quality}
\begin{figure}[!t]
    \centerline{\includegraphics[width=0.9\columnwidth]{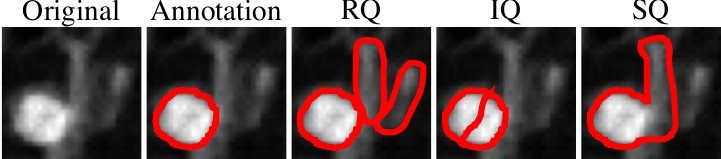}}
    \caption{Examples of idealized error types quantified by the \ac{ipq} metric.
    \ac{rq} detects hallucinations, \ac{iq} captures instance splitting, and \ac{sq} reflects oversegmentation.}
    \label{fig:seg_errors}
\end{figure}

\noindent
Segmentation performance is commonly evaluated using \ac{pq} \cite{kirillov_panoptic_2019}.
However, \ac{pq} fails to penalize the splitting of a correctly detected instance if each fragment exceeds the \ac{iou} threshold to be considered a \ac{tp}.
Such errors bias downstream measurements and must therefore be considered during optimization.
We introduce \acf{ipq}, an extension of \ac{pq}, as the objective function for segmentation optimization.
\ac{ipq} penalizes instance splitting and is composed of three factors, each capturing a distinct error type affecting interpretability.
The \acf{sq} $\in [0,1]$ factor evaluates over- and undersegmentation and is defined as
\begin{equation}
    \text{SQ} = 
    \frac{\displaystyle\sum_{(p, g) \in TP} 
    \text{IoU}\left(\displaystyle\bigcup_{p_i \in p} p_i,\, g\right)
    }{|TP|},
\end{equation}
where $TP$ is the set of \ac{tp} tuples $(p_i, g)$, with \textit{p} being a vector of predicted instances $p_i$ and the associated annotation \textit{g}.\\
The \acf{rq} $\in [0,1]$ factor penalizes hallucinations and omissions:
\begin{equation}
    \text{RQ} =
    \frac{|TP|}{|TP| + \frac{1}{2}|FP| + \frac{1}{2}|FN|}.
\end{equation}
The \acf{iq} $\in [0,1]$ factor penalizes instance splitting and is given by
\begin{equation}
    \text{IQ} =
    \frac{|G|}{\sum_{p \in P} \max(1, n_p - 1)},
\end{equation}
where $|G|$ is the number of annotations and $n_p$ is the number of predicted instances mapped to an annotation.
The combined metric is defined as
\begin{equation}
\text{IPQ} = 
(k_1 \text{SQ})\times(k_2 \text{RQ})\times(k_3 \text{IQ}),
\end{equation}
with weighting factors $k_i \in [0,1]$.
Fig.~\ref{fig:seg_errors} illustrates the error types detected by each factor.

\subsubsection{Data}
Manual annotation of 3D data is particularly challenging due to slice-based interfaces and computational constraints \cite{zheng_annotation_2020}.
We therefore recommend using synthetic 3D data to benchmark segmentation performance.
Domain gaps between synthetic and real data are addressed by combining data simulation with cycleGAN-based domain adaptation \cite{bruch_improving_2025, bohland_unsupervised_2023, bruch_synthesis_2023}.
This approach ensures complete and correct annotations for segmentation optimization.

\subsubsection{Parameter Space and Problem Definition}
To avoid computationally expensive retraining of pretrained segmentation models, segmentation optimization focuses on conceptual parameters, namely the choice of a model and postprocessing parameterization.
Postprocessing methods are selected to directly address the error types captured by \ac{ipq}, including morphological operations, instance merging, and instance splitting.
The optimization problem is formulated as
\begin{equation}
(\hat{m},\hat{\theta})
= \operatorname*{arg\,max}_{m \in M,\; \theta \in \Theta}
\sum_{x,y \in X} \mathrm{IPQ}\big(m(x),\, \theta, y\big),
\end{equation}
where $M=\{m_i(x)\}$ is a discrete set of segmentation models, $\Theta \subset \mathbb{R}^d$ denotes the continuous postprocessing parameter space, and $X$ is a synthetic 3D benchmark dataset of images $x$ and annotations $y$.
$\hat{m}$ and $\hat{\theta}$ are the optimal model and parameters obtained by the optimization process.

\subsubsection{Bayesian Optimization Setup}
Segmentation predictions are computed once per model and reused during optimization to avoid computationally expensive repetitions.
A Gaussian process surrogate models the postprocessing parameter space.
The acquisition function is \textit{expected improvement}, and optimization is terminated after 120 iterations.

\subsection{Classification Optimization}

\subsubsection{Objective and Data}
Classification optimization requires annotated data.
Instance segmentation using the optimized segmentation model predicts instances to enable efficient annotation without manual instance identification.
Additionally, a semi-supervised dataset is constructed using label spreading \cite{zhou_learning_2003}.
Feature vectors describing instance geometry are extracted, normalized, and reduced using principal component analysis.
Similarities between the feature vectors of the predicted instances are computed as the Euclidean distances using an RBF kernel in the reduced feature space to assign pseudo-annotations to all unannotated data of the semi-supervised dataset.
Datasets are split into training, validation, and test sets, with validation accuracy serving as the optimization objective.

\subsubsection{Parameter Space}
Classifier optimization targets conceptual design choices, including encoder and classifier head architectures, preprocessing, and pretraining strategies.
Multiple pretrained encoders of different sizes are included to balance under- and overfitting.
Classifier heads vary in number and size of linear layers to adapt the encoder feature space to the number of classes, and in the design of additional transformations.
Multiple architectures can be included in the \ac{bo} process to identify an optimal fit for the data.
Pretraining strategies determine whether encoder weights are frozen or retrained.
The \ac{pipe} chooses a pretraining strategy that reduces the risk of under- and overfitting to the data.
Preprocessing methods can incorporate a priori knowledge derived from segmentation predictions, such as instance position and geometry.
Each classifier configuration is trained using the Adam optimizer and cross-entropy loss, with data augmentation and early stopping.

\subsubsection{Bayesian Optimization Setup}
As each evaluation requires full classifier training, \ac{bo} is employed to efficiently explore the discrete parameter space.
A random forest surrogate is used, and \textit{expected improvement} serves as the acquisition function.

\section{Experiments}

\subsubsection{Overview}
We conduct four experiments to validate the \ac{pipe}.
For each optimization process, we propose explicit options for all optimized parameters. 
These options can be replaced by alternatives suited to a given application.
We further provide dedicated datasets for each \ac{bo} process.
For segmentation optimization, we construct a synthetic 3D benchmark per experiment and reduce the domain gap using a cycleGAN \cite{bruch_improving_2025}.
For classifier optimization, class annotations are obtained using the \ac{pipe}'s assisted instance-based annotation workflow.

\subsection{Data}

\subsubsection{Myotube Nuclei}
\begin{figure}[!t]
    \centering
    \includegraphics[width=0.99\linewidth]{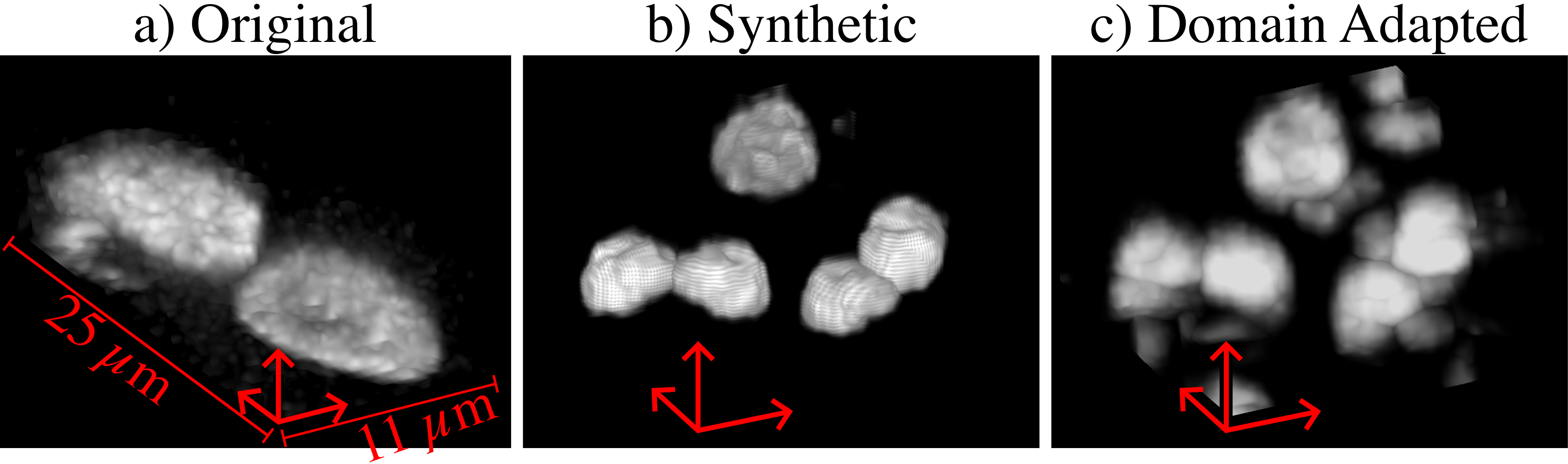}
    \caption{3D depiction of the synthetic segmentation benchmark. a) Original reference, b) synthetic image, and c) synthetic image after domain adaptation via cycleGAN.}
    \label{fig:data_synth_examples}
\end{figure}

The first experiment uses \textit{in vitro} Myotube cultures derived from human induced pluripotent stem cells, prepared following Couturier et al.~\cite{couturier_aberrant_2024}.
Three immunofluorescence channels are used for classification: DAPI, $\alpha$ sarcomeric actinin, and S100$\beta$.
We manually annotate 16 Schwann cell nuclei, 125 Myotube nuclei, 95 debris, and 148 other samples.
Because the Schwann cell class is small, 50\% of samples are assigned to the validation set.
The synthetic Myotube benchmark contains 111{,}458 nuclei across ten volumes of size $90 \times 1024 \times 1024$ pixels.
Fig.~\ref{fig:data_synth_examples} shows an exemplary 3D depiction of the real data and the synthetic benchmark.

\noindent

\subsubsection{Core-Shell}
The second experiment is conducted on cellular core-shell assembloids composed of transfected HEK293T cells, which comprise three cell classes.
A total of 360 cells are annotated using assisted annotation.
To emphasize validation-driven optimization, only 60 cells are used for training (48 train, 12 test), while 300 cells form an extensive validation set.
For segmentation optimization, we generate five synthetic volumes of size $70 \times 1024 \times 1024$ pixels containing 3{,}358 cells arranged inside the spheroid.

\subsubsection{Cell Tracking Challenge}
The 3D \ac{ctc} comprises ten datasets with annotations for segmentation and tracking \cite{maska_cell_2023}.
We include Fluo-C3DH-H157 (120 volumes, size $35 \times 832 \times 992$ pixels, 519 instances) and Fluo-C3DL-MDA231 (24 volumes, size $30 \times 512 \times 512$ pixels, 957 instances).
For these \ac{ctc} datasets, we do not synthesize data, as they already are established segmentation benchmarks.

\subsection{Parameters}

\subsubsection{Encoders}
The encoders in Table \ref{tab:encoders} are included in classifier optimization.
Five of the encoders are pretrained on ImageNet. 
The table shows their Acc@5, which indicates whether the correct class appears among the five most confident predictions on the ImageNet dataset \cite{russakovsky_imagenet_2015}.
Additionally, the CellposeSAM encoder is included.
\begin{table}[!h]
  \centering
  \caption{Comparison of the six encoders included.}
  \label{tab:encoders}
  \begin{tabular}{lcccc}
    \hline
    \textbf{Name} & \textbf{Abbrev.} & \textbf{Acc@5} & \textbf{Parameters (M)} \\
    \hline
    ResNet18 \cite{he_deep_2016} & RN18  & 89.08\% &	11.7 \\
    ResNet101 \cite{he_deep_2016} & RN101 & 93.55\% &	44.5 \\
    Swin V2 \cite{liu_swin_2022} & Swin  & 96.87\% &	87.9 \\
    EfficientNet V2 \cite{tan_efficientnetv2_2021} & EN    & 97.79\% &	118.5 \\
    ConvNeXt-L V1 \cite{liu_convnet_2022} & CN    & 96.98\% &	197.8 \\
    CellposeSAM \cite{pachitariu_cellpose-sam_2025} & CSAM  & 	-     &	305 \\
    \hline
  \end{tabular}
\end{table}
\subsubsection{Classifier Heads}
We include two standardized classifier heads derived from common practices \cite{wang_non-local_2018, liu_group_2021}: a \textit{slice classifier} and a \textit{volume classifier}.
The \textit{slice classifier} applies self-attention over a stack of averaged 2D slice representations to focus inter-slice dependency features.
The \textit{volume classifier} performs 3D convolutions over a 3D stack of 2D representations to efficiently encode volumetric features.
Architectural details are provided in the supplementary material.

\subsubsection{Pretraining Methods}
To balance over- and underfitting, we include three pretraining options.
(i) \textit{Fully-supervised pretraining}: ImageNet-pretrained encoder weights, and the classifier head are retained on manual labels with 80\% of encoder weights frozen.
(ii) \textit{Semi-supervised pretraining}: the encoder is first pretrained using the semi-supervised dataset, then retrained on manual labels with 80\% of weights frozen \cite{zhou_learning_2003}.
(iii) \textit{None}: the classifier is trained exclusively on the manually annotated dataset.

\subsubsection{Preprocessing Methods}
Segmentation predictions provide a priori knowledge about instance position and geometry.
We include two preprocessing methods to incorporate this knowledge.
The \textit{mask method} replaces the instance channel with the binary prediction to suppress neighboring instances and emphasize geometry, at the cost of attenuating surface features.
The \textit{distance method} retains surface features while encoding instance proximity by scaling intensities via a distance transform:
\begin{equation}
    I'(p_1) = I(x) \times \exp\Bigg(-\frac{1}{\sigma} \min_{p_2 \in P} \|p_1 - p_2\|_2 \Bigg),
\label{eq:distance}
\end{equation}
where \( I : \mathbb{N}^3 \to [0,1] \) is the intensity, 
\( P \subseteq \mathbb{N}^3 \) is the set of voxel positions inside the binary prediction, and $p_1$ denotes a voxel outside the mask.

\subsubsection{Segmentation Models}
Table \ref{tab:segm_models} summarizes the segmentation models included in segmentation optimization.
As the included StarDist model is a 2D model, we perform connectivity-based slice merging.
\begin{table}[h!]
    \centering
    \caption{Segmentation models included in all experiments.}
    \label{tab:segm_models}
    \begin{tabular}{ccc}
    \toprule
        \textbf{model} & \textbf{abbreviation} & \textbf{parameters} \\
        \hline
        StarDist 2D Versitile Fluo \cite{weigert_nuclei_2022, weigert_star-convex_2020, schmidt_cell_2018} & SD & 1.4B \\
        PlantSeg 3D Nuc Platinum \cite{yu_plantseg_2023} & PSP & 1B\\
        PlantSeg 3D Confocal UNet \cite{wolny_accurate_2020} & PSG & 4B\\
        CellposeSAM \cite{pachitariu_cellpose-sam_2025} & CSAM & 304B\\
        \bottomrule
    \end{tabular}
\end{table}

\subsubsection{Segmentation Postprocessing}
We optimize seven postprocessing parameters chosen to directly address the error types detected by \ac{ipq}.
Morphological operations (erosion/dilation and opening/closing) are applied to a binary mask that is then applied to the instance mask.
Instance splitting is performed using watershed, and instance merging is based on an adjacency graph with a distance metric combining contour continuity, connecting surface smoothness, and relative intersection area.
All optimized parameters are listed in Table \ref{tab:postProcessParams}.
\begin{table}[h]
\centering
\caption{Parameters included in the optimization of postprocessing methods.}
\begin{tabularx}{\linewidth}{P{0.04\linewidth}P{0.125\linewidth}P{0.1\linewidth}X}
\hline
 & \textbf{Parameter} & \textbf{Values} & \textbf{Description} \\
\hline
\multirow{2}{*}{\rotatebox{90}{Morphology }} 
& $\theta_{ED}$ & $[-10, 10]$ & Radius of the spherical structuring element for erosion (negative) or dilation (positive). \\
& $\theta_{CO}$ & $[-5, 5]$ & Radius of the spherical structuring element for an opening (negative) or closing (positive) operation. \\
\hline
& $\theta_{Mc}$ & $[0, 1]$ & Weight of contour continuity at intersections (from 1D profiles along the connecting axis) for the adjacency graph. \\
\multirow{2}{*}{\rotatebox{90}{Merging}} 
& $\theta_{Ms}$ & $[0, 1]$ & Weight of connecting-surface smoothness for the adjacency graph. \\
& $\theta_{Mr}$ & $[0, 1]$ & Weight of relative intersecting area (normalized by smaller-instance surface area) for the adjacency graph. \\
\hline
\multirow{2}{*}{\rotatebox{90}{Splitting }} 
& $\theta_{S\sigma}$ & $[0, 1]$ & Gaussian smoothing strength for watershed. \\
& $\theta_{St}$ & $[0, 1]$ & Peak density via distance-transform quantile threshold for watershed. \\
\hline
\end{tabularx}
\label{tab:postProcessParams}
\end{table}
All parameters of a category are set to zero to omit that postprocessing category entirely.

\section{Results \& Discussion}
\subsection{Segmentation}
\begin{figure}[t]
    \centering
    \includegraphics[width=0.99\linewidth]{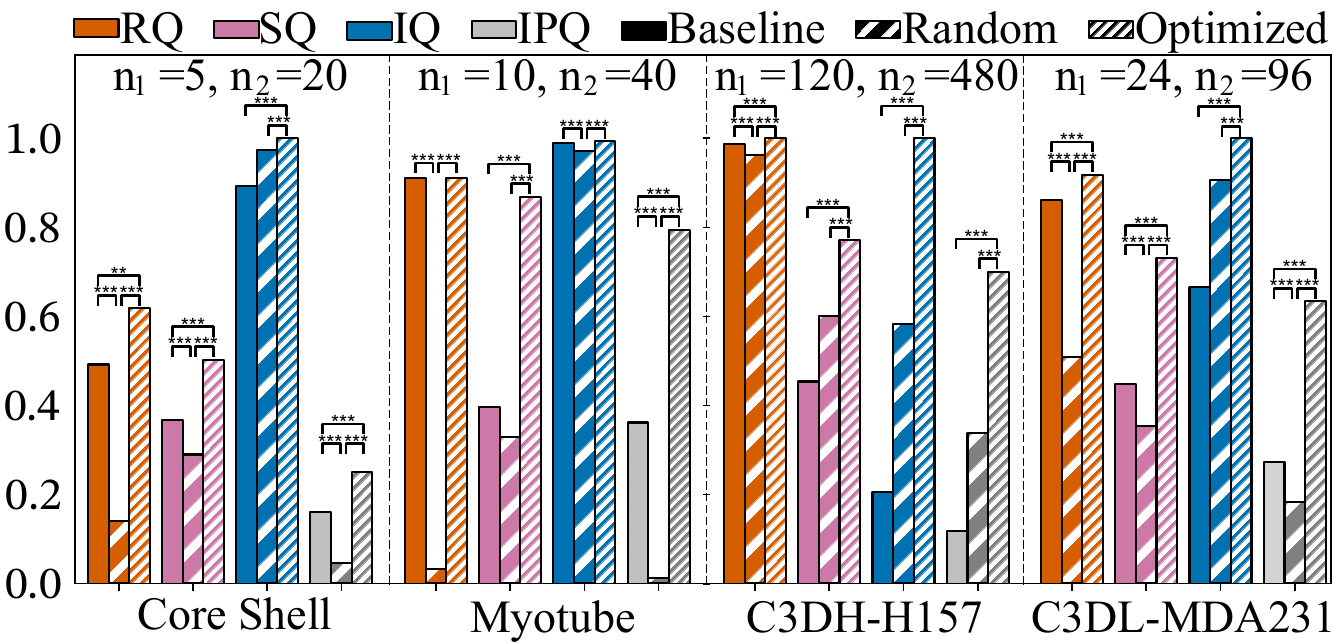}
    \caption{\ac{ipq} results of every experiment.
    Full bars present the baseline, thick-striped bars the random search, and thin-striped bars the model optimized by the \ac{pipe}.}
    \label{fig:per_model_experiment_IPQ}
\end{figure}

\begin{figure*}[t]
    \centering
    \includegraphics[width=0.99\linewidth]{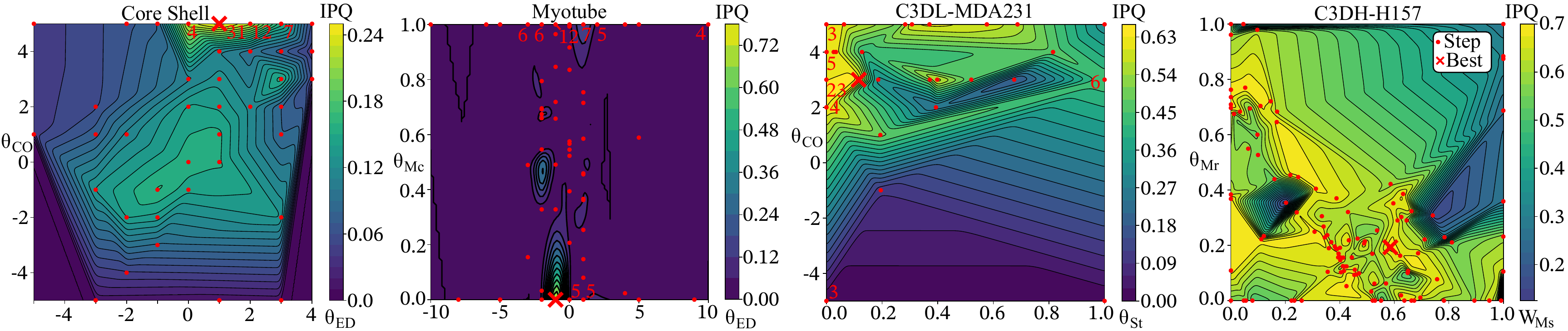}
    \caption{2D projection of a subset of the parameter space with resulting \ac{ipq} values. The dots denote exploration steps, and the cross denotes the found optimum.
    Red numbers next to dots indicate that a specific parameter value combination was evaluated multiple times.}
    \label{fig:bayesOpt}
\end{figure*}

To ensure data- and model-independent comparisons of the segmentation performance, we relate the results of each experiment to an experiment-specific baseline and to a random search with the same evaluation budget as the executed \ac{bo} process.
This baseline is computed as the mean \ac{ipq} of all models without any postprocessing, representing the performance of an unoptimized configuration.
Fig. \ref{fig:per_model_experiment_IPQ} shows the \ac{ipq} (grey) values as well as the \ac{rq} (orange), \ac{sq} (pink), and \ac{iq} (blue) factors of the baseline (filled bar), the random search (thick-striped bars), and the optimized segmentation model (thin-striped bar) for every experiment.
$n_1$ indicates the number of \ac{ipq} values for the random search and \ac{bo} process, and $n_2$ indicates the number of values for the baseline.
Statistical significance of every paired t-test is denoted by stars in the figure. 
\noindent
The optimized bars show significantly higher values than the baseline and the randomized search across all metrics in almost all experiments, indicating that the optimization process yields substantially better segmentation performance than randomly selecting a segmentation model and applying no postprocessing.
The low mean \ac{sq} value of 0.37 for the Core-Shell experiment suggests that the segmentation models cannot predict the correct instance areas from the synthetic data.
Because of this, morphological operators have great potential to improve segmentation performance, as they affect predicted instance sizes.
The significant improvement of 0.13 in \ac{sq} by the \ac{bo} process, and the decrease of 0.08 by the random search, show that the \ac{bo} process was successful in identifying parameters to enhance the predicted areas, whereas the random search was not. \\
The Myotube experiment shows no \ac{rq} and \ac{iq} improvement by the \ac{bo} process over the baseline, but a 0.48 improvement in the \ac{sq} metric.
This indicates that the pretrained segmentation models already successfully identify all instances, but they predict them either too large or too small, highlighting the importance of morphological operators for the Myotube data. 
Every metric decreases significantly with the random search.
This shows that the parameter space is too complex for a random search to find a suitable solution in a limited number of iterations.\\
For the \ac{ctc} experiments, we observe a particularly strong improvement of 0.79 and 0.33 in the \ac{iq} metric, while \ac{rq} only improves by 0.01 and 0.06.
Before optimization, the models tested here confidently find almost all instances, as indicated by the high baseline \ac{rq} of 0.99 and 0.86, but they repeatedly split individual instances in their prediction, as indicated by the low baseline \ac{iq} of 0.21 and 0.67, rendering a good parameterization of merging algorithms imperative.
While the random search does identify a parameterization that improves \ac{iq} by 0.38 and 0.24, the improvement is significantly lower than that of the \ac{bo} process, and it entails a decrease of \ac{rq} by 0.02 and 0.35.
The \ac{bo} process thus not only alleviates the weakness of the baseline more efficently than a random search, but also improves upon the pretrained models' strengths.\\
Next, we demonstrate that identifying an optimal parameterization is difficult to do manually in this complex search space, and that a \ac{bo} process can determine effective settings.
Fig. \ref{fig:bayesOpt} shows exemplary contour plots of two postprocessing parameters that are particularly important for each experiment. 
The plots show the resulting \ac{ipq} and include only evaluations that use the model achieving the highest \ac{ipq}. 
The first contour plot shows the Core-Shell experiment search space, including both morphological operators.
As mentioned, they are especially important to this experiment.
In the plot, two local optima are visible, one being $\theta_{ED}=0, \theta_{CO}=0$, which is also the starting point for the \ac{bo} process.
While approaches like Coordinate Descent will get stuck in local optima, as this plot shows, the \ac{bo} process is able to find a superior configuration in this case, indicating that the implementation of a \ac{bo} process is worthwhile.
Next, the $\theta_{ED}$-$\theta_{Mc}$ parameter space of the Myotube data experiment is displayed.
The mean baseline \ac{rq} and \ac{iq} are 0.92 and 0.99, but the \ac{sq} is only 0.4.
Because the pretrained model already identifies almost all instances correctly, instance merging was effectively ruled out by the \ac{bo} process, and the contour plot only shows \ac{ipq} values higher than 0.2 at the optimum of $\theta_{Mc} = 0.0$.
At the global optimum, all merging and splitting parameters are zero.
This shows that the \ac{bo} process can decide to omit postprocessing methods completely if required.
Along the $\theta_{Mc} = 0$ axis, $\theta_{ED} = -1$ shows a substantial improvement with an \ac{ipq} value of 0.8.
This indicates that the pretrained segmentation model was biased toward predicting slightly larger instances, which the \ac{bo} process identified and fixed.
A table of the optimal parameterization with the resulting \ac{ipq} for every experiment and segmentation model is given in the supplementary material.\\ 
The third contour plot is of the C3DH-H157 data.
As mentioned before, the parameterization of the merging algorithm is especially important for this experiment, thus the merging parameters $\theta_{Ms}$ and $\theta_{Mr}$ are displayed.
The shape of the search space is very complex due to the high number of other parameters influencing the performance. 
Still, an optimum is found in a densely explored region of the search space. 
This demonstrates the necessity of an efficient optimization of postprocessing parameterization, as manually finding a good solution would be challenging here.
Similarly, the second \ac{ctc} experiment, the C3DL-MDA231 data, shows a splitting algorithm parameter and the Closing/Opening parameter.
The contour plot shows only a few exploration steps that include a high value for the splitting algorithm.
Because they lead to low \ac{ipq} values, the majority of the parameter configurations tested, as well as the optimum, lie in the range $\theta_{St} < 0.2$.
The second parameter on display shows that higher $\theta_{CO}$ values lead to higher \ac{ipq} values, presumably because the segmentation models introduce gaps between the predicted instances they split an individual annotation instance with, which are closed by the closing operator.
Search space regions of negative $\theta_{CO}$ values are thus not extensively explored.

\subsection{Classification}
\begin{figure*}
    \centering
    \includegraphics[width=0.99\linewidth]{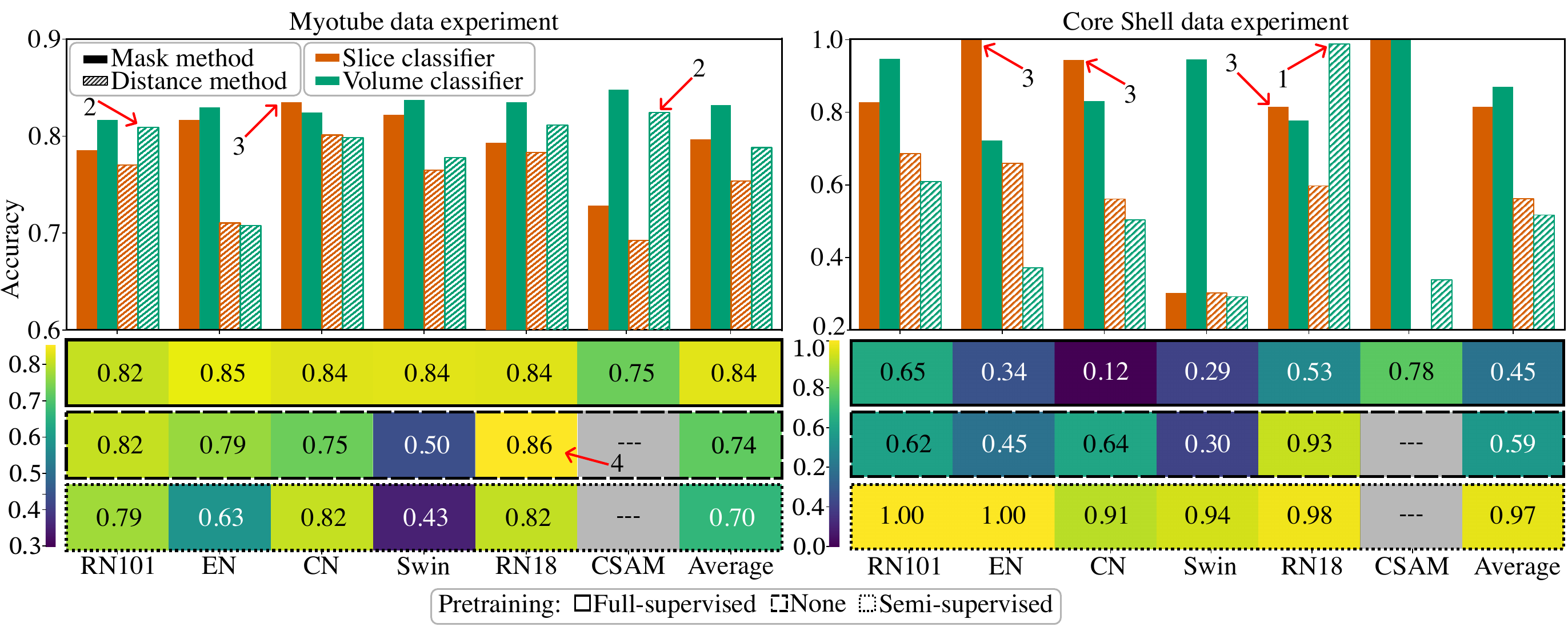}
    \caption{Results of the classification optimization processes per encoder, classifier head, pretraining method, and preprocessing method with four highlights:
    1) On average worst combination performs best, 2) second to worst combination performs best, 3) \textit{slice classifier} outperforms \textit{volume classifier}, and 4) best pretraining strategy is outperformed.}
    \label{fig:results_classifier}
\end{figure*}

Fig. \ref{fig:results_classifier} shows an overview of the validation accuracies of the classifiers trained in the classifier optimization process.
A full table including every validation accuracy is given in the supplementary material.
On the left, the results of the Myotube data experiment are given, and on the right, the Core-Shell data experiment results are displayed.
Both bar graphs show the mean accuracy of each classifier using a specific encoder and a specific classifier head-preprocessing method combination.
Below that, heatmaps visualize the mean validation accuracies achieved by the classifier for every encoder–pretraining combination.
\noindent
The bar graphs reveal that the \textit{mask method} is, on average, superior to the \textit{distance method} across both experiments, with accuracy differences of 4\% and 29\% respectively.
For the Myotube data experiment, the \textit{volume classifier} is, on average, 3\% better than the \textit{slice classifier}.
Validation accuracies range from 12\% to 100\%.
This indicates that the design choices matter and that the process is able to identify methods that outperform others. \\
We observe that some combinations outperform combinations with higher average accuracies.
For example, the \textit{volume classifier} using the \textit{distance method} with the RN18 encoder achieves an accuracy of 98.7\% on the Core-Shell data (1), even though the combination of \textit{volume classifier} and \textit{distance method} is the worst on average (51\%).
Similarly, the \textit{volume classifier}-\textit{distance method} combination has a 10\% and 2\% higher accuracy than \textit{slice classifier}-\textit{mask method} for the CSAM and RN101 encoders, respectively, on the Myotube data (2).
The \textit{slice classifier}-\textit{mask method} combination in turn often outperforms the on average best combination across both experiments (3).
Out of the six encoders per experiment, only two show the same interaction pattern between design parameters as observed in the average of the encoders.
For the remaining encoders, the relative performance of each classifier head-preprocessing method combination differs.
This variation of relative combination performance suggests that the average performance is not representative of individual combinations and, therefore, that a more in-depth search is necessary to identify optimal combinations. \\
For the Myotube data experiment, \textit{fully-supervised pretraining} yields the best results on average with 84\%. 
In contrast, for the Core-Shell data, it is better to pretrain the encoders on a semi-supervised dataset (97\%).
This shows that the results of the optimization process cannot be generalized to find the perfect classifier design for every dataset. \\
The overall best classifier for the Myotube data is the smallest model, an RN18 encoder with no pretraining (4), likely because the \ac{cnn} architecture is sufficiently simple, and the encoder is small enough to generalize well on the limited dataset.
We further prove the hypothesis, that smaller, simpler encoders generalize better by computing the Pearson correlation between the number of parameters of the \ac{cnn} encoders and the validation accuracy, obtaining a correlation coefficient of $r=-0.97, p=0.027$.  
Table \ref{tab:encoders_durations} shows the inference times of every encoder to process one image.
We observe that, for example, the RN18 encoder achieves the same maximum accuracy on the Core-Shell dataset as the CSAM encoder while processing about five times faster.
Similarly, on the Myotube data, the RN18 encoder (86\%) and the EN encoder (85\%) achieve maximum accuracies on par with the overall maximum, achieved by the CSAM encoder (87\%) with a much shorter processing duration.
The optimization process thus also reveals trade-offs of processing time and accuracy, enabling classifier design based on application requirements.
\begin{table}[!h]
  \centering
  \caption{Comparison of inference durations per encoder.}
  \label{tab:encoders_durations}
  \begin{tabular}{lcccc}
    \hline
    \textbf{Name} & \textbf{Duration one image (s)}\\
    \hline
    ResNet18 	    & 0.42 \\
    ResNet101 	    & 0.47 \\
    Swin V2	        & 0.61 \\
    EfficientNet V2  & 0.59 \\
    ConvNeXt-L V1    & 0.73 \\
    CellposeSAM	    & 2.23 \\
    \hline
  \end{tabular}
\end{table}
\noindent  
\appendices
\section*{Acknowledgments}
The authors declare no conflict of interest.
\section*{References and Footnotes}
\printbibliography
\begin{acronym}
    \acro{pipe}[3D-AOP]{3D data Analysis Optimization Pipeline}
    \acro{bo}[BO]{Bayesian Optimization}
    \acro{ipq}[IPQ]{Injective Panoptic Quality}
    \acro{pq}[PQ]{Panoptic Quality}
    \acro{sq}[SQ]{Segmentation Quality}
    \acro{rq}[RQ]{Recognition Quality}
    \acro{iq}[IQ]{Injective Quality}
    \acro{tp}[TP]{True Positive}
    \acro{fp}[FP]{False Positive}
    \acro{fn}[FN]{False Negative}
    \acro{iou}[IoU]{Intersection over Union}
    \acro{cnn}[CNN]{Convolutional Neural Networks}
    \acro{ctc}[CTC]{Cell Tracking Challenge}
\end{acronym}
\newpage
\section*{Supplementary material}
\subsection{Classifier heads}\label{app:classifiers}
\begin{figure}[h!]
    \centering
    \includegraphics[width=0.85\linewidth]{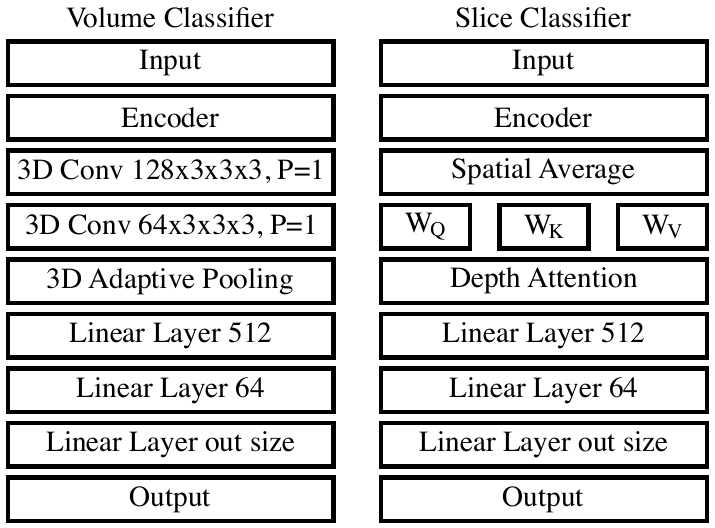}
    \caption{Architectures of the classifier heads. 
    The \textit{slice classifier} performs a self-attention mechanism over the encoders' representations of the 2D slices of a volume.
    The \textit{volume classifier} applies 3D convolutions on the stacked 2D representations.}
    \label{fig:classifiers}
\end{figure}

The architecture of the classifier heads included in the optimization process are given in Fig. \ref{fig:classifiers}.
\subsection{Segmentation results}\label{app:SegRes}
Table \ref{tab:MyoRes} shows the best parameterization of all experiments and segmentation models with the resulting \ac{ipq}. 
\begin{table}[h!]
\centering
\caption{\ac{ipq} results of every segmentation model with its optimal postprocessing parameterization per experiment.}
\begin{tabular}{|c|c|cc|ccc|cc|c|}
\hline
\multicolumn{1}{|c}{} &
\multicolumn{1}{c|}{} &
\multicolumn{2}{c|}{Morphological Operators} &
\multicolumn{3}{c|}{Merging Parameters} &
\multicolumn{2}{c|}{Splitting Parameters} &
\multicolumn{1}{c|}{Result} \\
\hline
Experiment & Model &  $\theta_{ED}$ & $\theta_{CO}$ & $\theta_{Mc}$ & $\theta_{Mr}$ & $\theta_{Ms}$ & $\theta_{S\sigma}$ & $\theta_{St}$ & \ac{ipq} \\
\hline
\multirow{4}{*}{C3DH-H157} & CSAM & -4 & 4 & 0.073 & 0.584 & 0.194 & 0.915 & 0.000 & 0.699 \\
 & SD & 5 & 3 & 0.472 & 0.125 & 0.977 & 0.317 & 0.587 & 0.244 \\
 & pSG & 1 & 2 & 0.862 & 0.321 & 0.882 & 0.695 & 0.823 & 0.237 \\
 & pSP & 10 & 2 & 0.306 & 0.479 & 0.254 & 0.744 & 0.496 & 0.488 \\
\hline
\multirow{4}{*}{C3DL-MDA231} & CSAM & 1 & 2 & 1.000 & 1.000 & 0.000 & 0.746 & 0.000 & 0.637 \\
 & SD & 1 & 0 & 1.000 & 1.000 & 0.000 & 1.000 & 0.000 & 0.561 \\
 & pSG & -2 & 3 & 0.000 & 0.240 & 1.000 & 0.174 & 0.548 & 0.364 \\
 & pSP & -2 & 2 & 0.718 & 0.000 & 0.000 & 1.000 & 0.000 & 0.223 \\
\hline
\multirow{4}{*}{Core Shell} & CSAM & 3 & 3 & 0.769 & 0.537 & 0.017 & 0.033 & 0.514 & 0.112 \\
 & SD & 1 & 5 & 1.000 & 0.050 & 1.000 & 0.393 & 0.000 & 0.251 \\
 & pSG & -2 & 5 & 0.019 & 0.000 & 0.964 & 0.597 & 0.000 & 0.164 \\
 & pSP & 1 & 5 & 1.000 & 0.040 & 1.000 & 0.396 & 0.000 & 0.168 \\
\hline
\multirow{4}{*}{Myotube} & CSAM & -1 & 0 & 0.000 & 0.000 & 0.000 & 0.000 & 0.000 & 0.799 \\
 & SD & 0 & 1 & 1.000 & 0.000 & 1.000 & 1.000 & 0.000 & 0.012 \\
 & pSG & 0 & -5 & 0.499 & 0.991 & 0.001 & 0.576 & 0.256 & 0.217 \\
 & pSP & 3 & 0 & 0.009 & 0.550 & 0.988 & 0.917 & 0.074 & 0.092 \\
\hline
\end{tabular}
\label{tab:MyoRes}
\end{table}

\subsection{Classificaiton optimization results}
Table \ref{tab:cls_results_sideways} shows all validation accuracies collected in the classification optimization 
\begin{sidewaystable*}[t!]
\centering
\caption{Observed accuracies per experiment, encoder, pretraining, preprocessing, and classifier head.}
\footnotesize
\begin{tabular}{llcccccccccccc}
\toprule
 &  & \multicolumn{12}{c}{Pretraining} \\
\cmidrule(lr){3-14}
 &  & \multicolumn{4}{c}{Fully-supervised} &
       \multicolumn{4}{c}{None} &
       \multicolumn{4}{c}{Semi-supervised} \\
\cmidrule(lr){3-6}\cmidrule(lr){7-10}\cmidrule(lr){11-14}
 &  & \multicolumn{2}{c}{Mask} & \multicolumn{2}{c}{Distance} &
       \multicolumn{2}{c}{Mask} & \multicolumn{2}{c}{Distance} &
       \multicolumn{2}{c}{Mask} & \multicolumn{2}{c}{Distance} \\
\cmidrule(lr){3-4}\cmidrule(lr){5-6}
\cmidrule(lr){7-8}\cmidrule(lr){9-10}
\cmidrule(lr){11-12}\cmidrule(lr){13-14}
Experiment & Encoder
& Slice & Volume & Slice & Volume
& Slice & Volume & Slice & Volume
& Slice & Volume & Slice & Volume \\
\midrule
Myotube & RN18 & 0.79427 & 0.83594 & 0.78385 & 0.8125 & - & 0.85938 & - & - & - & 0.81510 & - & - \\
Myotube & RN101 & 0.78646 & 0.81771 & 0.77083 & 0.8099 & - & 0.81771 & - & - & - & 0.79167 & - & - \\
Myotube & Swin & 0.82292 & 0.83854 & 0.76562 & 0.77865 & - & 0.5 & - & - & - & 0.42969 & - & - \\
Myotube & EN & 0.81771 & 0.83073 & 0.71094 & 0.70833 & - & 0.78646 & - & - & - & 0.63021 & - & - \\
Myotube & CN & 0.83594 & 0.82552 & 0.80208 & 0.79948 & - & 0.7526 & - & - & - & 0.81771 & - & - \\
Myotube & CSAM & 0.72917 & 0.84896 & 0.69271 & 0.82552 & - & - & - & - & - & - & - & - \\
\midrule
Core-Shell & RN18 & 0.652317881 & - & - & - & - & 0.890728477 & 0.367549669 & 0.605960265 & 1 & 1 & 1 & - \\
Core-Shell & RN101 & - & - & 0.314569536 & 0.367549669 & - & 0.450331126 & - & - & 1 & 0.990066225 & 1 & - \\
Core-Shell & Swin & - & - & 0.11589404 & - & 0.887417219 & 0.658940397 & - & 0.367549669 & 0.996688742 & 1 & 1 & 0.632450331 \\
Core-Shell & EN & 0.298013245 & - & - & 0.28807947 & - & - & 0.298013245 & - & - & 0.943708609 & - & - \\
Core-Shell & CN & 0.645695364 & 0.738410596 & 0.205298013 & - & - & 0.811258278 & 0.983443709 & 0.986754967 & 0.98013245 & - & - & - \\
Core-Shell & CSAM & 1 & 1 & - & 0.334437086 & - & - & - & - & - & - & - & - \\
\bottomrule
\end{tabular}%
\label{tab:cls_results_sideways}
\end{sidewaystable*}

\end{document}